# Energy saving in smart homes based on consumer behavior: A case study

Michael Zehnder, Holger Wache, Hans-Friedrich Witschel
Institute of Business Information Systems
University of Applied Sciences and Arts Northwestern Switzerland
FHNW Olten, Switzerland
mzehnder@gmx.net
{holger.wache; hansfriedrich.witschel}@fhnw.ch

Danilo Zanatta, Miguel Rodriguez
Research and Development
digitalSTROM AG
Zurich, Switzerland
{danilo.zanatta; miguel.rodriguez}@digitalstrom.com

*Abstract*—This paper presents a case study of a recommender system that can be used to save energy in smart homes without lowering the comfort of the inhabitants. We present an algorithm that mines consumer behavior data only and applies machine learning to suggest actions for inhabitants to reduce the energy consumption of their homes. The system looks for frequent and periodic patterns in the event data provided by the digitalSTROM home automation system. These patterns are converted into association rules, prioritized and compared with the current behavior of the inhabitants. If the system detects opportunities to save energy without decreasing the comfort level, it sends a recommendation to the inhabitants.

The system was implemented and deployed to a set of test homes. The test participants were able to rate the impact of the recommendations on their comfort. This feedback was used to adjust the system parameters and make it more accurate during a second test phase. The historical data set provided by digitalSTROM contained 33 homes with 3521 devices and over 4 million events. The system produced 160 recommendations on the first phase and 120 on the second phase. The ratio of useful recommendations was close to 10%. We found out that a recommender system that uses an algorithm that mines patterns based on their confidence, independent of their frequency and periodicity, might achieve better results and a higher acceptance by users.

*Keywords—smart cities; smart homes; energy saving; recommender system; association rules; unsupervised learning; internet of things; IoT*

## I. INTRODUCTION

King defines a smart home as a "dwelling incorporating a communications network that connects the key electrical appliances and services" and which can be "remotely controlled, monitored or accessed" [1]. Blumendorf enhances the "smart" part of the definition with a "home which does act in a smart way, is a system which is autonomously operating, based on artificial intelligence" [2]. But there would be many ways to measure the smartness of such a home. Harper acknowledged that "smart house technologies that most people are pleased with are connected with saving energy or money" [3]. As previous studies show, energy savings can only be achieved by involving the inhabitants [4].

User comfort is a soft characteristic that cannot be objectively measured as it is the case with savings in energy consumption. In this paper we propose a recommender system, which relies on the opinion of the users to approve a recommendation. We evaluate the smartness of the recommender system designed in this project by the energy savings created without reduction in comfort. A long-term goal is for the algorithm to learn which kind of patterns would be likely to be accepted by the user because it does not impact their comfort.

This paper is organized as follows. Chapter II provides some background about smart homes and unsupervised learning. Chapter III describes the pattern-mining algorithm used and Chapter IV describes the design of the recommendation system. The results from the evaluation of the recommender system in *real* life are presented in Chapter V, while conclusions and future work are described in Chapter VI.

## II. BACKGROUND

### A. Smart home technology

digitalSTROM products provide connectivity to electrical devices in the home over the existing power cables. This includes every lamp, light switch, blinds and any plugged in device. This network of devices is connected through a server mounted in the electrical cabinet to a local area network. The result is a network of connected devices, bringing the internet of things (IoT) to each home. digitalSTROM components are based on a high volt IC in a small size module. Each digitalSTROM module can switch, dim, measure electricity and communicate its status. The products are available through Europe with its larger installed base in Germany and Switzerland.

A digitalSTROM system is based on concentrators that reside in the electrical distribution panel, acting as power meters for the individual distribution circuits and communicating with individual nodes installed within a home over differently modulated up and downstream channels [5]. The system includes a Linux server application with a JSON API. Moreover, real-time data from test homes are collected by a logging system, parsed and stores in a database, being available for processing. From this logging system, historical data can easily be obtained and processed.

Please note that for this approach no motion detectors or other location sensors where used. Only activities like turning on (or off) light, TV etc. are considered.

*B. Unsupervised learning and mining for patterns*

Many supervised activity recognition methods for smart homes have been published in the literature in the last years (see e.g. [6]). Supervised learning is used in methods like decision trees [7], Markov models [8] and dynamic Bayes networks [9]. Although these classifiers rely on conditional independence of the events, the classifiers achieve good accuracy with large amounts of training data [10].

However, these methods have the major disadvantage that the training data needs manual annotation. This manual annotation is a very tedious and time-consuming task and limits the scalability of the system. Furthermore, the training data can only be used for the household for which they were made for [11].

Due to the limitation listed above, in this work, an unsupervised approach for pattern mining is used. In the unsupervised approach, the user is not required to scan his data for activities and the classifier is able to find the patterns autonomously.

## III. PATTERN MINING

The recommender system proposed in this work is based on the mining of patterns from historical event data from a home in order to produce recommendations for its users. This chapter describes the choice of the pattern mining algorithm and the adaptations required in order that it can be used in the smart home context.

*A. Data analysis*

Finding frequent and periodic patterns in event data is usually referred to as "discovering frequent episodes in sequences" [12], "sequence mining" [6, 13] or "activity recognition" [9, 14, 15, 16, 17]. The starting point for this research was the Apriori algorithm [23], which was later implemented and enhanced by other projects. The Apriori algorithm in its original form is not able to mine periodic patterns in sequential data and was therefore improved by different people for the use in smart home automation systems, for example for the CASAS System [10]. In this approach, the algorithm starts by mining patterns with a minimum length of two activities, which is extended until the algorithm is not able to find frequent patterns anymore.

*B. Data set – the events*

The basic structure of the house is the hierarchical model shown by Fig. 1. The root node represents a home, one household. A home has several dSMs (dSM stands for "digitalSTROM Meter"), one per electric circuit in the home, and several zones, where each zone represents a physical room in the house.

Each zone contains one or many scenes, where each scene represents a pre-defined configuration of the devices, e.g. each light dimmed to a pre-defined state, shades closed and television turned-on.

Each dSM controls the devices that are connected to its electrical circuit. Power measurements are recorded per dSM. Events are either inhabitant generated button clicks or sensor events, such as temperature sensors.

The historical training dataset used contains 33 homes with 3521 devices, which are related to 4,331,443 events and 6829 unique scenes. These events extend over a period between 08/12/2002 and 25/06/2014.

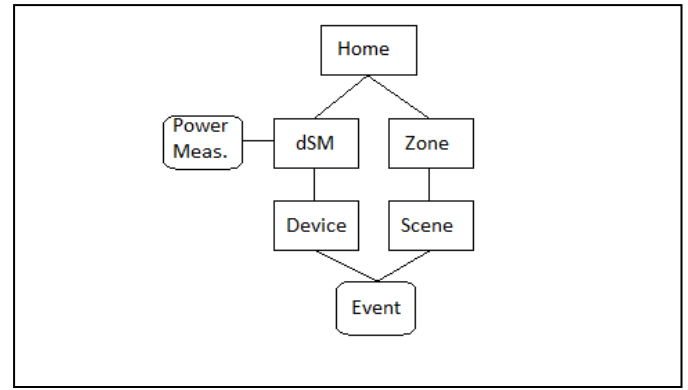

Fig. 1. Hierarchy of components in a digitalSTROM smart home

*C. Definition and identification of patterns that would reduce energy usage*

Analysis of the data revealed that the power measurement data is not recorded as precisely as necessary for obtaining information about the power consumption of a single device. In contrast, the event data is well suited and therefore eligible to find possible actions.

We defined rules to identify relevant energy saving actions from the underlying data. We considered the following type of events as energy saving, and therefore as an action for the system:

- Absent (a predefined scene which turns off all devices when leaving the house)
- Dim (scenes to dim devices)
- Off (scenes to turn off devices)
- Sleep (scenes to turn off devices before going to sleep)
- Standby

In the training data this rule results in a list of 20 actions, which are called by 1,283,756 unique events.

*D. Identification of relevant types of behavior patterns*

Because not all frequent or periodic patterns result in energy savings, we defined some characteristics to identify the relevant behavior patterns.

To ensure that a relevant pattern can be used to suggest actions, it must be composed of two main components:

(1) A relevant pattern must contain at least one *action* to lower energy usage

(2) The pattern must consist of *normal events* (not actions), which serve as condition to suggest the action at the right time.

Suppose that such patterns occur regularly, they reflect the normal behavior of the inhabitants. The main idea now is to discover situations where the user forgot the action. Therefore the recommender would suggest the user to perform this action to lower energy usage.

Because a one-event condition is insufficient as evidence to suggest an action, the overall length of a relevant pattern must be at least three events (one action and two normal events). Therefore, a relevant pattern can be defined as a pattern that is longer than two events and contains at least one action to lower energy usage.

Because a relevant pattern consists of normal events, which represents the condition, and an action, this pattern can be interpreted using an association rule. An association rule is an implication of the form

$$X \rightarrow Y, \text{ where } X, Y \subset I, \text{ and } X \cap Y = \emptyset,$$

where X is a sequence of normal events, Y is a single action, and I is the set of all possible events. The association rule above states that when X occurs, Y occurs with certain probability [22]. This rule is also depicted in Fig. 2.

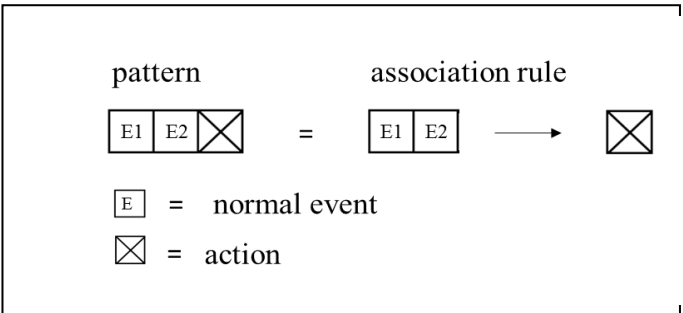

Fig. 2. Sequence of events and the association rule

To summarize, the types of behavior patterns that are required to recommend some action must fulfill the following criteria:

- The pattern must occur frequently and/or periodically in the data
- The pattern must be relevant for action prediction and contain at least one action to lower energy usage
- The pattern must have a minimum length of three events, including normal events and actions

*E. Evaluation of patern-mining algorithms*

Three established frequent sequential pattern-mining algorithms PrefixSpan, BIDE+ and GapBIDE as well as their adaptations were evaluated by Schweizer in [19] for the usage on smart home event data. They were benchmarked against each other and against a self-developed algorithm named WSDD. All four algorithms mined a reasonable amount of frequent sequential patterns. Under the following parameters, nearly thousand patterns were found in the historical test data:

- pattern length: between 3 and 7 events
- minimum support: between 0.01 and 0.001
- overlapping patterns allowed
- wildcarding disabled

The traditional frequent sequential pattern mining algorithms like PrefixSpan, BIDE+ or GapBIDE require pre- and post-processing to be used for mining smart home event data. Furthermore, if different minimum and maximum lengths of patterns shall be mined, those algorithms need to run multiple times to report the correct support count.

The run times of the different algorithms showed large deviations, as shown in Fig. 3. While BIDE+ needed the longest to mine the same patterns as the other three algorithms, both GapBIDE and PrefixSpan run significantly faster. However, they were all outperformed by the WSDD algorithm.

The relevant patterns mined in the underlying data showed lengths of 3 to 6 events. Whereas the minimum of three events is a predefined parameter, the maximum was not defined. However, testing on the underlying data showed that no relevant pattern was longer than six events.

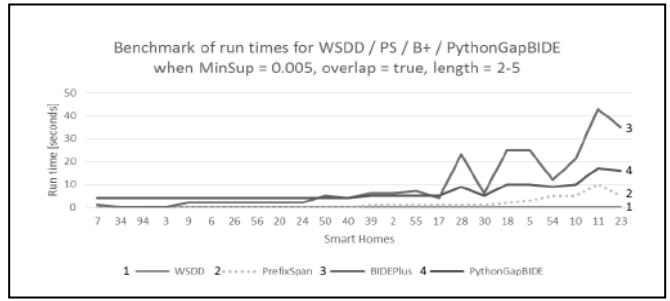

Fig. 3. Benchmark of run times for data mining algorithms

The following table presents three examples of relevant patterns mined in the underlying event data by the relevant pattern-mining algorithm. The grey background of the cell indicates the action, which must be part of a relevant pattern.

| Pattern | Events | | |
|---|---|---|---|
| | *1* | *2* | *3* |
| 1 | Turn off light in kitchen | Turn on light in stairs | Motion detector garage |
| 2 | Turn on light in laundry room | Turn off light in laundry room | Turn off light in basement |
| 3 | Turn on light outside of keller | Turn on lamp in keller | Turn off lamp in keller |

It should be noted that the order of the events/actions is not important. As the examples above show, an action can appear at the first position. The two next events for the condition might appear after the action but can be used to discover the missing action.

IV. DESIGN OF THE RECOMMENDER SYSTEM

*A. Architecture*

The architecture of the recommender system developed in this project is shown in Fig. 4 and can be divided in three main parts:

- The storage of the association rules
- The event stream of the current behavior data inside the smart home
- The matching algorithm

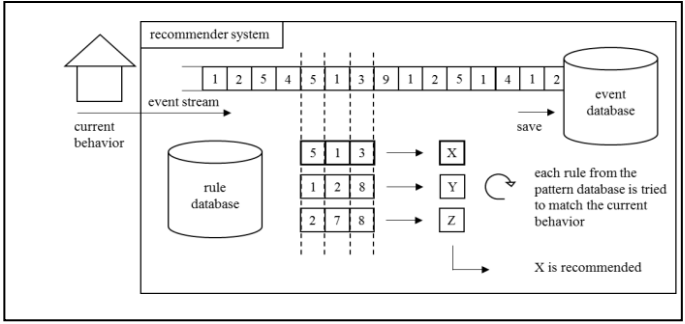

Fig. 4. Recommender system architecture

The **rule database** stores the association rules, which were obtained from the mined relevant patterns. The **event stream** contains the real-time events from the smart home, ordered by time of their occurrence.

The **matching algorithm** is the core component of the recommender system. It matches the rules and the event stream. The most common existing rule matching algorithm is RETE by Forgy [21]. We used a deterministic finite state machine (FSM) approach as depicted in Fig. 5, which reflects the order of the events better than RETE. A new instance of the FSM is created for each new event in the stream. If there is no matching in the first attempt, the instance is removed from memory. If the condition did match and the next event is not the action itself, the machine sends a recommendation.

The design of the recommender system allows more than one rule to be matched at the same time. In order to avoid multiple conflicting recommendations, we propose to weight the rules and use the weights as a prioritization criterion. Furthermore, the prioritization criterion is also used to exclude rules under a certain threshold, i.e. rules that are *weakly* matched. The criteria where defined as follows:

- Confidence of the rule
- Support of the pattern
- The length of the pattern
- The position of the action
- Date when the pattern was mined

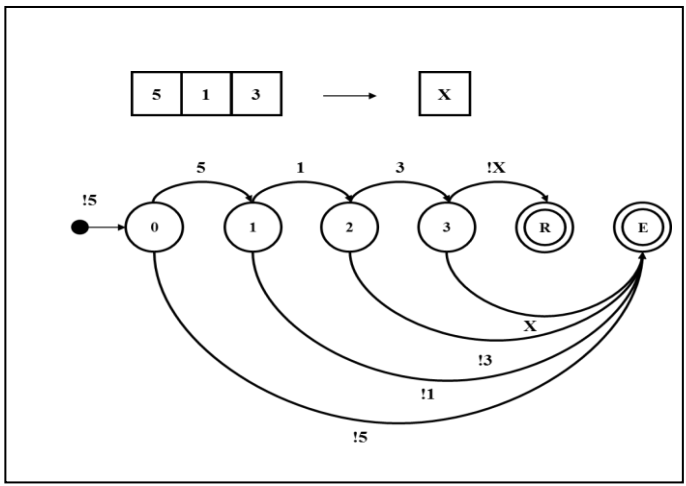

Fig. 5. Example of a matching rule in the algorithm (FSM)

## B. Confidence

The confidence of a rule denotes how often the action Y appears in patterns that contain the sequence of events X (condition). If a rule has a confidence of 100%, no occurrence of the pattern without the action was mined. The confidence is calculated using the support (count) of the pattern. The confidence is thus expressed as

$$\text{confidence}(\boxed{{}^{1\,2\,3\,4}_{\boxtimes\square\square\square}}) = \frac{\text{support}(\boxed{{}^{1\,2\,3\,4}_{\boxtimes\square\square\square}})}{\text{support}(\boxed{{}^{1\,2\,3\,4}_{\boxtimes\square\square\square}}) + \text{support}(\boxed{{}^{1\,\,3\,4}_{\square\,\square\square}})}$$

where $\boxtimes$ denotes the action, $\square$ denotes a normal event, $\boxed{{}^{1\,2\,3\,4}_{\boxtimes\square\square\square}}$ denotes the pattern containing the action and $\boxed{{}^{1\,\,3\,4}_{\square\,\square\square}}$ denotes the pattern without the action.

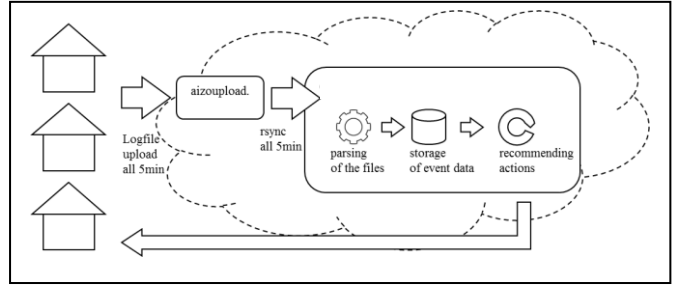

Fig. 6. Recommender deployment view

## C. Implementation

The recommender system was implemented on a cloud based Microsoft® Azure VM (Virtual Machine). The VM was set up with Ubuntu 14.04. The deployment view of the VM is shown in Fig. 6.

The smart homes event data is parsed from the log files of the digitalSTROM system. The files are uploaded by a script installed on the digitalSTROM infrastructure in the houses and made accessible for this project on a file server. The files are copied every 5 minutes by remote synch to the VM where the recommender system is running. They are parsed by a Python script and stored in a MySQL database.

## V. RESULTS

The evaluation was conducted in households with inhabitants producing real event data. The test households were equipped with the smart home automation system digitalSTROM. Overall, 33 households were considered for the evaluation. The historical event data of the homes was mined by the recommender system in advance. According to the number of relevant patterns found in the houses, they were ranked and the owners of the 15 most promising households where requested to take part in the evaluation. From these, 8 houses agreed to participate in the evaluation of this work including both single- and multi-inhabitant houses. Recommendations were sent per SMS to the mobile devices of the inhabitants. An example of such recommendation is shown in Fig. 7.

We ran the evaluation in two phases, which are described in the following sections.

```
    Hi Michael Zehnder,
    I would recommend to turn-off device Bed
side lamp (on 2014-11-16 23:19:16).
    Is this recommendation useful?
    Yes:
http://snurl.com/29fgj3d?r=2663&c=4f=1
    No:
http://snurl.com/29fgj3d?r=2663&c=4f=0
```

Fig. 7. Recommendation SMS

TABLE I. KEY RESULTS OF EVALUATION

| Parameter | Phase | |
|---|---|---|
| | 1 | 2 |
| # days evaluated | 14 | 34 |
| Recommendations sent | 160 | 120 |
| Answered recommendations | 76 | 55 |
| Voted useful | 7 | 5 |
| Voted not useful | 69 | 50 |
| Ratio useful/answered | 9.21% | 9.10% |
| Number of active rules | 54 | 46 |
| Number of rules that resulted in recommendations | 23 | 17 |
| Number of rules with 10 negative feedbacks | 5 | 3 |

### A. Phase 1

The aim of the first phase was to provide a large basis of data for evaluation and further improvement of the system. The analysis of the data collected during phase 1 should help to improve the recommender system in terms of decreasing the negatively rated recommendations in phase 2, while holding the positives at the highest amount possible. The results for phase 1 are summarized in TABLE I.

After running the evaluation for phase 1 for 2 weeks, the inhabitants were interviewed and their feedback revealed two main findings:

- Absent scene should not be recommended as action – it doesn't make sense to recommend a user to *leave home*
- The low response rate of 47.5% was caused by ambiguous recommendations – some users have similar or equal names for different devices inside their household. Consequently, a recommendation to turn off the "spot light" could apply for more than one device and the users could not decide if the recommendation was useful and ended up not responding at all

We did a regression analysis of the results using the *weighted feedback* as the dependent variable and the following prioritization served as explanatory variables:

- The length of the pattern
- The position of the action
- Support of the pattern
- Confidence of the rule

Because repeating the pattern mining on the event data barely produced new patterns, the date when the rule was mined was ignored as explanatory variable.

Patterns with high confidence and high pattern-length tend to receive better feedback than the other patterns. On the other hand, support and position of the action did not show any significance to describe the feedback a rule. It is worth to notice that support was the major attribute for mining frequent patterns. Fig. 8 shows the regression based on confidence.

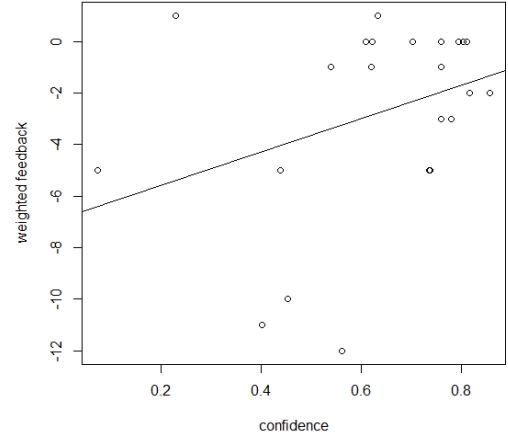

Fig. 8. Scatter plot of confidence with line of best fit

### B. Phase 2

The aim of the second phase of the evaluation was to increase the ratio of answered recommendations and the ratio of useful recommendations compared to the first phase. The analysis of the data collected during phase 1 was used to adapt the system, which should lower negative rated and unanswered recommendations. The prototype was improved in the following points:

- All rules which were excluded during phase 1, with 10 negative feedbacks in a row, were removed from phase two. The feedback count from phase 1 was reset for all patterns
- All rules which recommended the action "absent", where excluded from the second phase
- To reduce the problem of the low response rate caused by ambiguous recommendations, we enriched the text with the name of the room of the device. Since no manual matching was necessary, this feature is still in the scope of this work
- As result of the regression analysis, confidence and pattern-length of each rule where multiplied with their estimate to calculate a coefficient which gives indication about the *usefulness* of a rule. A threshold is defined and 19 rules out of 54 with a coefficient below this value where excluded from the second phase (35 rules remained).

The results from phase 2 are summarized in TABLE I. as well. The results show a similar ratio of useful recommendations as in phase 1, as well as a similar response rate (45.8%). However, a significant improvement can be observed in the number of recommendations sent: 0.44 recommendations/day/home in phase 2 *versus* 1.43 recommendations/day/home in phase 1. Note that, for the same ratio of useful recommendations, a lower number of recommendations per day per home means less *noise* for the user and a better comfort level.

## VI. CONCLUSION AND FUTURE WORK

In this work, we presented a case study of a system that generates recommendations to save energy in smart homes without reducing the comfort of the inhabitants. The results show that such a system works in *real life* and achieved a ratio of useful recommendations of about 10%, while sending 0.44 recommendations/day/home.

Several points of improvement were identified during the evaluation phases of this work. A follow-up research project is already ongoing and will build upon the findings of this work. The following ideas for further research materialized during the design, implementation or evaluation of the recommender system:

- Using confidence and pattern length instead of support or periodicity as criteria for the mining algorithm, resulting in more and better patterns
- The time between two events (or the action) is considered neither by the mining algorithm nor by the recommender system. Using this information will improve the accuracy of the suggestions made by the system
- Other attributes could be introduced to decide if a rule is relevant or not. Such attributes might be:
    - Time of day when the pattern occurs most
    - Weekday when the pattern occurs most
    - Season when the pattern occurs most
- The recommender should learn from the feedback of the inhabitants in order to prioritize the rules, instead of just excluding a rule after 10 negative feedbacks in a row
- Test other machine learning algorithms and frameworks such as Torch and Coffe